\documentclass[journal]{IEEEtran}
\usepackage{cite}
\usepackage[pdftex]{graphicx}
\usepackage{amsmath}
\usepackage{algorithmic}
\ifCLASSOPTIONcompsoc
 \usepackage[caption=false,font=normalsize,labelfont=sf,textfont=sf]{subfig}
\else
 \usepackage[caption=false,font=footnotesize]{subfig}
\fi

\usepackage{amssymb,amsfonts}
\usepackage{xcolor}
\usepackage{booktabs}
\usepackage{algorithm}
\usepackage[hidelinks]{hyperref}
\usepackage{orcidlink}
\newcommand{\myhyperlabel}[1]{\hypertarget{#1}{}}
\newcommand{\figref}[2][]{\hyperlink{#2}{\figurename~\ref*{fig:#2}#1}}
\newcounter{algocf}
\newcommand{\algcaption}[1]{%
  \begingroup 
  \addtocounter{algocf}{1}
  \renewcommand{\figurename}{Algorithm} 
  \renewcommand{\thefigure}{\thealgocf} 
  \caption{#1}
  \addtocounter{figure}{-1}
  \endgroup 
}

\begin{document}

\title{DE-PADA: Personalized Augmentation and Domain Adaptation for ECG Biometrics Across Physiological States}

\author{Amro~{Abu~Saleh}\textsuperscript{1}\,\orcidlink{0009-0004-3759-2270},
        Elliot~Sprecher\textsuperscript{2}\,\orcidlink{0000-0001-8564-1090},
        Kfir~Y.~Levy\textsuperscript{1}\,\orcidlink{0000-0003-1236-2626},
        and~Daniel~H.~Lange\textsuperscript{1}\,\orcidlink{0009-0008-2650-6153}%
\thanks{

\textsuperscript{1} Department of Electrical and Computer Engineering, Technion Israel Institute of Technology}%
\thanks{\textsuperscript{2} Independent Researcher.}
\thanks{Corresponding author: samro@campus.technion.ac.il}
}

\maketitle

\begin{abstract}
Electrocardiogram (ECG)-based biometrics offer a promising method for user identification, combining intrinsic liveness detection with morphological uniqueness. However, elevated heart rates introduce significant physiological variability, posing challenges to pattern recognition systems and leading to a notable performance gap between resting and post-exercise conditions. Addressing this gap is critical for advancing ECG-based biometric systems for real-world applications.
We propose DE-PADA, a Dual Expert model with Personalized Augmentation and Domain Adaptation, designed to enhance robustness across diverse physiological states. The model is trained primarily on resting-state data from the evaluation dataset, without direct exposure to their exercise data. To address variability, DE-PADA incorporates ECG-specific innovations, including heartbeat segmentation into the PQRS interval, known for its relative temporal consistency, and the heart rate-sensitive ST interval, enabling targeted feature extraction tailored to each region's unique characteristics. Personalized augmentation simulates subject-specific T-wave variability across heart rates using individual T-wave peak predictions to adapt augmentation ranges. Domain adaptation further improves generalization by leveraging auxiliary data from supplementary subjects used exclusively for training, including both resting and exercise conditions.
Experiments on the University of Toronto ECG Database (UofTDB) demonstrate the model’s effectiveness.
DE-PADA achieves relative improvements in post-exercise identification rates of 26.75\% in the initial recovery phase and 11.72\% in the late recovery phase, while maintaining a 98.12\% identification rate in the sitting position.
These results highlight DE-PADA’s ability to address intra-subject variability and enhance the robustness of ECG-based biometric systems across diverse physiological states.
\end{abstract}


\section{Introduction}

\IEEEPARstart{E}{lectrocardiogram} (ECG) signals have been widely recognized as a highly promising biometric modality due to their hidden nature and unique liveness detection capabilities, which provide a significant advantage over traditional biometric systems such as fingerprint or facial recognition \cite{Uwaechia2021}. These traditional systems are particularly vulnerable to sophisticated spoofing attacks, as they lack intrinsic liveness verification, which is crucial for ensuring that the biometric trait being analyzed belongs to a living individual.

The physiological and anatomical differences in heart structure, reflected in distinctive ECG waveform morphology, make ECG signals inherently unique and strong candidates for biometric applications \cite{interindividualECG, 8392675}. Their non-invasive nature and low-cost acquisition further reinforce their potential for widespread use.

Despite these advantages, ECG-based biometric systems face significant challenges in physiologically variable conditions.
While prior studies demonstrate high accuracy when users are in a controlled, resting state sitting position, their performance degrades in other conditions, with a significant degradation in non-resting states, such as post-exercise \cite{Wahabi2014, Hwang2021, Jyotishi2022}.
This decline in performance is primarily due to variability in ECG waveforms caused by numerous factors, including body posture, physical activity, and emotional state, all of which can induce substantial changes in the signal's morphology. These variations present considerable challenges for pattern recognition algorithms, which are expected to accurately identify individuals despite these fluctuations.
Pathoumvanh et al. \cite{Pathoumvanh} demonstrated that the performance decline is strongly correlated with heart rate changes, showing that even a moderate 40\% increase in heart rate due to physical activity can lead to a performance decrease of over 30\%.

In this work we address the challenge of user identification under elevated heart rates, 
using post-exercise conditions as a representative scenario. Such conditions are essential to consider for the development of practical ECG-based biometric systems, as they reflect the physiological changes that naturally occur during everyday activities.

The variability in ECG signals is mainly observed in the duration of heartbeat intervals and their amplitudes \cite{sornmo2005bioelectrical}, with the interval duration being inversely correlated with heart rate.
Various approaches have been proposed to address this variability, leveraging both traditional signal processing techniques and advanced machine learning models, particularly deep learning. Several studies have adjusted the duration of specific heartbeat intervals to mitigate intra-subject variability \cite{Hwang2021, heartID, Arteaga-Falconi2016, Choi2020, 5580317}. In contrast, deep learning-based methods often utilize data augmentation strategies to increase the variability in training data, enhancing model robustness to high variability \cite{8219706, Um_2017, Kim2022}.
However, most of these studies have focused on scenarios with limited data availability in resting conditions or relied on private databases that include exercise conditions.

The main contributions of this work are as follows.

\begin{itemize}

    \item Introduction of a Dual-Expert Model: We propose a novel Dual-Expert model that incorporates prior knowledge on ECG into the deep neural network architecture, enhancing its ability to handle the complexities of ECG signal variability.
    
    \item Personalized Augmentation: We introduce a subject-specific augmentation technique that mitigates the performance degradation observed in resting states with conventional augmentation methods, while also improving computational efficiency.
    
    \item Domain Adaptation for various conditions: We propose a domain adaptation variant to enhance the classifier’s generalization across diverse physiological conditions.

\end{itemize}

The combination of the contributions above form The Dual Expert with Personalized Augmentation and Domain Adaptation (DE-PADA) model, which demonstrates notable performance improvements on the University of Toronto ECG Database (UofTDB).
DE-PADA achieves a relative increase of 26.75\% in identification rates in the post-exercise initial recovery phase and 11.72\% in the late recovery phase, while maintaining a 98.12\% identification rate in the sitting position.
The effectiveness of each contribution is further detailed in Section~\ref{sec:ablation_study}, highlighting their integral role in enhancing the model's performance across diverse physiological states.
\section{Related Work}

There are two primary approaches in the literature to tackle heart rate variability. The first approach is heartbeat interval normalization, which aims to correct the duration of different heartbeat intervals to a canonical state, thereby reducing variations caused by heart rate changes. The second approach involves data augmentation that aims to generate synthetic data to enrich the training set, often in conjunction with deep learning techniques to enhance feature extraction. 

\subsection{Heartbeat Interval Normalization}
Various normalization methods targeting the duration of specific ECG intervals have been proposed to reduce the effect of heart rate changes.
Fatemian et al. \cite{heartID} resampled the T-wave section to align it with the standard healthy duration under resting conditions, focusing on stabilizing the most variable segment of the ECG.
Arteaga-Falconi et al. \cite{Arteaga-Falconi2016} introduced a linear normalization method that normalizes the temporal features of the ECG relative to the total heartbeat duration, improving authentication reliability despite physiological fluctuations.
Choi et al. \cite{Choi2020} proposed a fusion normalization approach that combines time-domain and frequency-domain normalization techniques to improve the alignment and consistency of ECG signals. This method linearly interpolated the P and T waves from post-exercise recordings, extending these segments to match estimated pre-exercise durations.
Hwang et al. \cite{Hwang2021} conducted regression analyses on an auxiliary dataset to establish relationships between PR, QRS, ST, and TP intervals and heart rate.
Based on these relationships, canonical interval durations were calculated at a baseline heart rate of 70 bpm.
Normalizing each interval to this canonical state demonstrated improved performance across diverse conditions, including post-exercise states.

While these approaches show improved performance, they apply uniform adjustments to interval durations across all subjects, disregarding the individual variability in the relationship between heart rate and interval durations \cite{Malik2002}. This uniformity assumption may lead to inaccuracies in normalization, reducing alignment with the actual physiological characteristics of individual subjects.

\subsection{Data Augmentation}
Unlike normalization methods, data augmentation techniques aim to increase system robustness by simulating intra-subject variability during model training.
Random alterations such as permutation, cropping, noise addition, and scaling of different ECG intervals have been proposed to diversify training datasets \cite{8219706, Um_2017}.
However, these generic augmentations do not account for changes induced by heart rate variability.
Kim et al. \cite{Kim2022} introduced a physiology-based augmentation technique targeting the ST interval. Using Hodges’ QT interval correction formula \cite{hodges1983bazett}, this method linearly resampled the ST interval over a range of durations for each subject, mimicking physiological variability.
While this approach improved model performance on resting-state data and addressed limitations associated with a small training dataset, the method did not account for individual variability among subjects, potentially resulting in inaccurate approximations.
\section{Methodology}
\subsection{Database Overview}
We use the UofT database (UofTDB) \cite{Wahabi2014}, which was collected in an off-the-person setting and includes recordings in various positions. This database is designed to facilitate evaluation in practical environments and under a wide range of challenging conditions that biometric systems might encounter. UofTDB contains recordings from participants across up to six different sessions and five conditions: sitting, standing, supine, tripod, and exercise. The distribution of subjects across these conditions is summarized in Table~\ref{tab:subject_distribution}.

\begin{table}[!t]
    \centering
    \caption{Subject Count by Condition and Recording Session in UofT Database}
    \label{tab:subject_distribution}
    \begin{tabular}{ccccccc}
        \toprule
        \textbf{Session} & \textbf{Sit} & \textbf{Stand} & \textbf{Exercise} & \textbf{Supine} & \textbf{Tripod} \\
        \midrule
        S1 & 1012 & 0   & 0  & 0  & 0   \\
        S2 & 72   & 72  & 0  & 0  & 0   \\
        S3 & 76   & 5   & 71 & 0  & 0   \\
        S4 & 63   & 0   & 0  & 0  & 0   \\
        S5 & 0    & 0   & 0  & 63 & 63  \\
        S6 & 65   & 65  & 0  & 0  & 0   \\
        \bottomrule
    \end{tabular}
\end{table}

To effectively analyze the impact of heart rate dynamics on system performance, we aim to minimize the influence of other sources of variability by integrating them into the training data.
To achieve this, recordings from multiple sessions are included to capture day-to-day variations.
Additionally, the training dataset incorporates recordings from both sitting and standing positions, representing the range of realistic and practical postures for ECG acquisition.
This approach ensures a more robust evaluation while accounting for common physiological and environmental factors.
The \textit{Target} set, which serves as the primary dataset for training and evaluation, includes subjects recorded in all five positions.
In addition, subjects with exercise recordings but missing data in other positions, which were excluded from the Target set, form an \textit{Auxiliary} set of 15 subjects, which is utilized in later stages.

\subsection{Signal Preprocessing}
\begin{figure}[!t]
\myhyperlabel{preprocess_overview}
\centering
\includegraphics[width=2.5in]{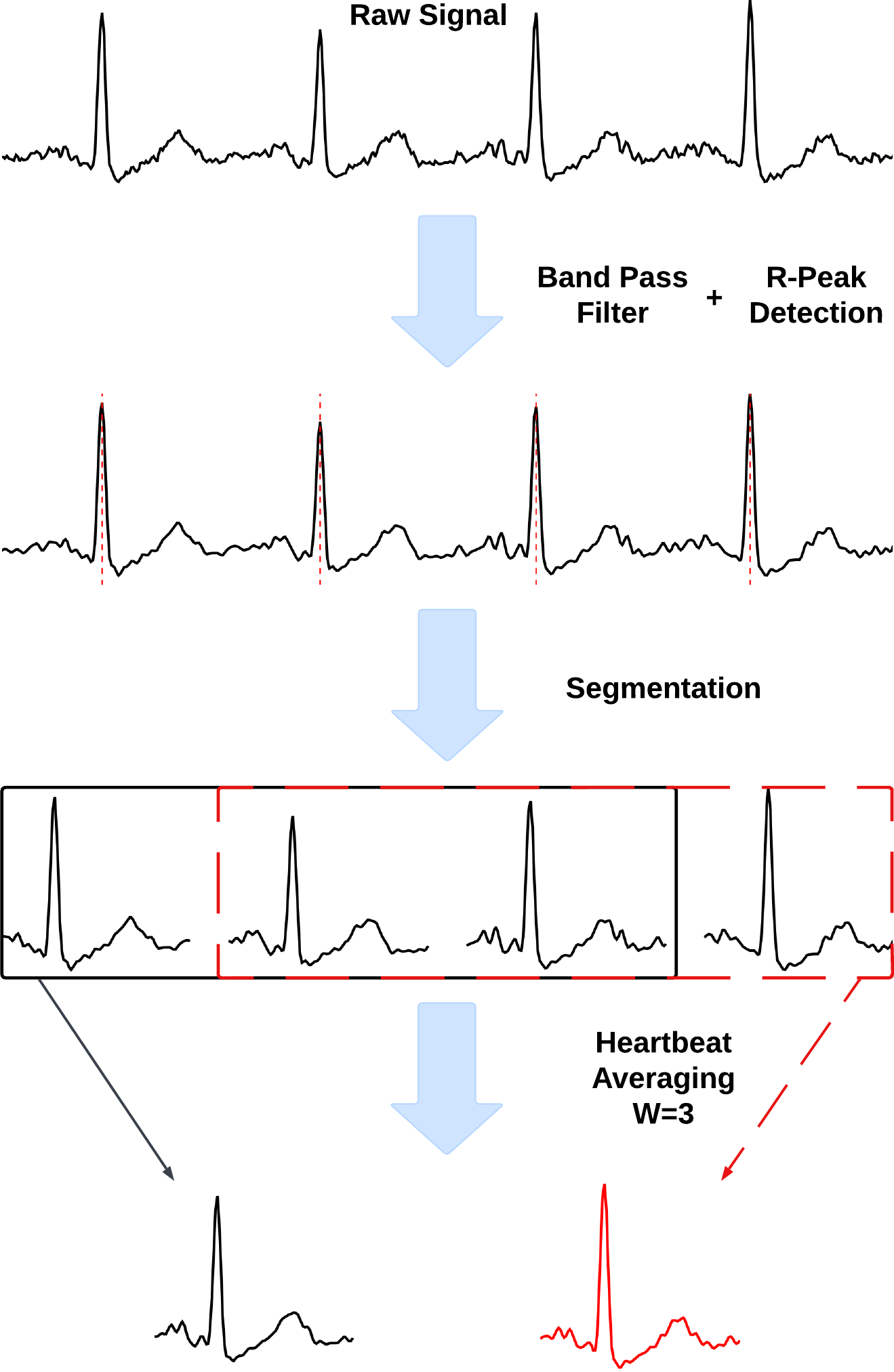}
\caption{Overview of the preprocessing flow, with an average size of $W=3$ used for illustration.}
\label{fig:preprocess_overview}
\end{figure}

The preprocessing workflow is presented in \figref{preprocess_overview}.
\subsubsection{Signal Filtering}
Following previous studies \cite{Jyotishi2022, 8392675} and the database authors \cite{Wahabi2014}, the raw ECG signals are filtered using a 4th order Butterworth band-pass filter with cutoff frequencies of 0.5 Hz and 40 Hz to remove baseline wander, power line interference, and other artifacts.
\subsubsection{R-peak Detection}
Following filtering, R-peaks are detected using the neurokit2\cite{Makowski2021neurokit} library and the biopeaks \cite{Brammer2020} method. These tools were chosen for their robustness and accuracy in detecting peaks in noisy signals compared to the commonly used Pan-Tompkins method \cite{4122029}, which is critical for precise heartbeat segmentation.
However, due to noise and artifacts in off-the-person databases, sharp spikes are sometimes incorrectly identified as R-peaks. Inter-quartile range (IQR) method \cite{Dekking2005} was used to remove heartbeats with R-peak amplitude falling below the lower threshold $(Q_1 - 1.5 * IQR)$ or above the upper threshold $(Q_3 + 1.5 * IQR)$, with the quantiles calculated for each recording separately.
\subsubsection{Heartbeat Segmentation}
The signals are segmented into 550 milliseconds (ms) heartbeats, with 175 ms before and 375 ms after the R-peak. These segmentation parameters balance capturing key waveform features while minimizing overlap between heartbeats, particularly in high heart rate signals.

Off-the-person ECG datasets often suffer from a low signal-to-noise ratio (SNR) due to various factors such as movement artifacts and environmental noise. To mitigate this issue, consecutive heartbeats are averaged using a sliding window technique with a window size $W$. This averaging process improves the SNR by approximately $\sqrt{W}$, although it reduces the number of available heartbeats in each recording.

The window size $W$ is a hyper-parameter that can be adjusted based on the signal quality of the ECG device, providing flexibility in preprocessing both the training and testing sets. This adaptability enables the use of single heartbeats, which is particularly important in scenarios that demand minimal acquisition time.

\subsubsection{T-wave Peak Detection}
For each segmented heartbeat, we locate the T-wave peak (T-peak) by finding the point of maximum amplitude within the last 300 ms of the heartbeat. To enhance the accuracy of T-peak detection, an outlier removal process based on Z-scores is applied to the T-peak locations in the time domain.
Z-score values are calculated separately for the entire rest-state training data and the exercise data.
As the exercise data in the Target set is reserved exclusively for evaluation, the Auxiliary set is used to derive thresholds for exercise samples.
We set the thresholds to be three standard deviations from the mean location for each group, effectively filtering out erroneous T-peak detections and enhancing the reliability of the T-peak identification process.

\subsection{Personalized Augmentation}

To increase the variability of the training data and compensate for the lack of exercise or elevated heart rate samples, we aim to generate synthetic data that simulates heartbeats across a range of heart rates.
Specifically, to simulate heartbeats at different heart rates, we augment the ST interval in the time domain, as this is the most variable part due to heart rate changes.

Building on prior work in ST interval augmentation \cite{Kim2022} and normalization \cite{Hwang2021}, we observed that regressing the location of the T-wave peak (T-peak) against the heart rate exhibits the expected linear relationship. This aligns with some of the QT corrections used in medical diagnosis \cite{hodges1983bazett, SAGIE1992797} and supports the findings in \cite{Hwang2021}.
While the linear relationship holds across the general population, our analysis reveals that when analyzing each individual separately, the slope of the linear curve differs among subjects, providing a more precise reflection of each subject's unique characteristics.

Therefore, we calculate the linear slope for each subject individually, based on their training data in sit and stand positions. This approach allows us to predict the individual range of T-peak locations for each subject. Subsequently, we perform ST interval augmentation within these personalized ranges, rather than using a global fit or applying a fixed range across all subjects.

Given that the number of sitting recordings is significantly larger than the standing recordings, we calculate both balanced and unbalanced linear fits for each individual.

The unbalanced linear fit assigns a uniform weighting to all the data points, while the balanced linear fit distributes the weights so that the overall weight of the sit data points equals the overall weight of the stand data points.

Heart rate variability is low for some subjects even when considering both sit and stand positions, which can cause one of the linear fits to deviate.
To address this issue, we calculate the minimal T-peak value for the global, balanced, and unbalanced linear fits. The closest value to the global fit among the balanced and unbalanced fits is then selected as the lower T-peak limit, as detailed in \hyperlink{individual_augmentation}{Algorithm 1}.
The upper T-peak limit corresponds to the lowest heart rates. Since resting data is included in the training set, there is no need to generate augmented data up to the subjects' maximum range. Instead, we aim to generate data only in the range where it is lacking; therefore, the upper limit is set as the median of the T-peak locations from the standing position in the training data.

\begin{figure}[!t]
\hrule\noindent
\myhyperlabel{individual_augmentation}
\begin{algorithmic}[1]
    \STATE \COMMENT{--- Initialization ---}
    \STATE $HR_{\text{limit}} \gets 140$ \COMMENT{Upper bound heart rate for augmentation}
    \STATE $T_{g_{\text{min}}} \gets 29$ \COMMENT{$T_{peak}$ location at $HR_{\text{limit}}$ for global fit}
    \STATE $T_{p_{\text{min}}} \gets 25$ \COMMENT{Physiological minimal $T_{peak}$ location}
    \STATE $S_k$: Signals for each individual $k$

    \STATE \COMMENT{--- Main Algorithm ---}
    \FOR{each individual $k \in \{0, \ldots, N-1\}$}
        \STATE Compute $T$-wave location at $HR_{\text{limit}}$ using individual linear fit \label{lst:line:pa_start}
        \STATE \hspace{0.5em} $T_b[k] \gets \text{Balanced fit result}$
        \STATE \hspace{0.5em} $T_{ub}[k] \gets \text{Unbalanced fit result}$

        \IF{$(T_b[k] - T_{g_{\text{min}}})^2 \leq (T_{ub}[k] - T_{g_{\text{min}}})^2$}
            \STATE $T_{peak_{\text{min}}}[k] \gets T_b[k]$
        \ELSE
            \STATE $T_{peak_{\text{min}}}[k] \gets T_{ub}[k]$
        \ENDIF
        \STATE $T_{peak_{\text{min}}}[k] \gets \max(T_{peak_{\text{min}}}[k], T_{p_{\text{min}}})$
        \STATE $T_{peak_{\text{max}}}[k] \gets \text{median}(\{T_{peak} \mid T_{peak} \in \text{individual } k\})$ \label{lst:line:pa_end}

        \FOR{$s \in S_k$}
            \FOR{$T_{new} \in [T_{peak_{\text{min}}}[k], T_{peak_{\text{max}}}[k]]$}
                \STATE $s^{PQRS} \gets \text{PQRS interval of }s$
                \STATE $s^{ST} \gets \text{ST interval of }s$
                \STATE $s^{ST}_{res} \gets \text{Resample } s^{ST}$ so that its T-peak is aligned with  $T_{new}$
                \STATE $s_{res} \gets \text{Concatenate }[s^{PQRS}, s^{ST}_{res}]$
            \ENDFOR
        \ENDFOR
    \ENDFOR
\end{algorithmic}
\noindent\hrule
\algcaption{Personalized Augmentation Algorithm}
\label{alg:individual_augmentation}
\end{figure}
\subsection{Standard CNN Architecture}
We employ a Convolutional Neural Network (CNN) as our feature extraction backbone, followed by fully connected layers for classification. The architecture is adapted from \cite{9211012}, with modifications to accommodate a smaller input size and reduce the number of Max-pooling layers. Additionally, batch normalization layers were added to stabilize and speed up training, as illustrated in \figref[a]{standard_dual_models}. After the convolutional layers, the feature maps are flattened and passed into a Multi-Layer Perceptron (MLP) with a single hidden layer, for classification. The output dimension of the MLP corresponds to the number of classes.

\subsection{Dual Expert Model}

Deep learning models generally perform well when test data matches the distribution of the training data, assuming sufficient data is available.
However, generalizing to exercise conditions without including exercise data in the training set introduces additional challenges.
Several previous studies have split ECG heartbeats into segments, addressing some of the segments separately \cite{Chun2016, Saechia}, but these studies relied on traditional processing methods. 
However, studies employing deep learning models typically rely on the model to extract the necessary features, thereby avoiding manual segmentation of the heartbeat.

\begin{figure}[!t]
\myhyperlabel{pqrs_st}
\centerline{\includegraphics[width=\columnwidth]{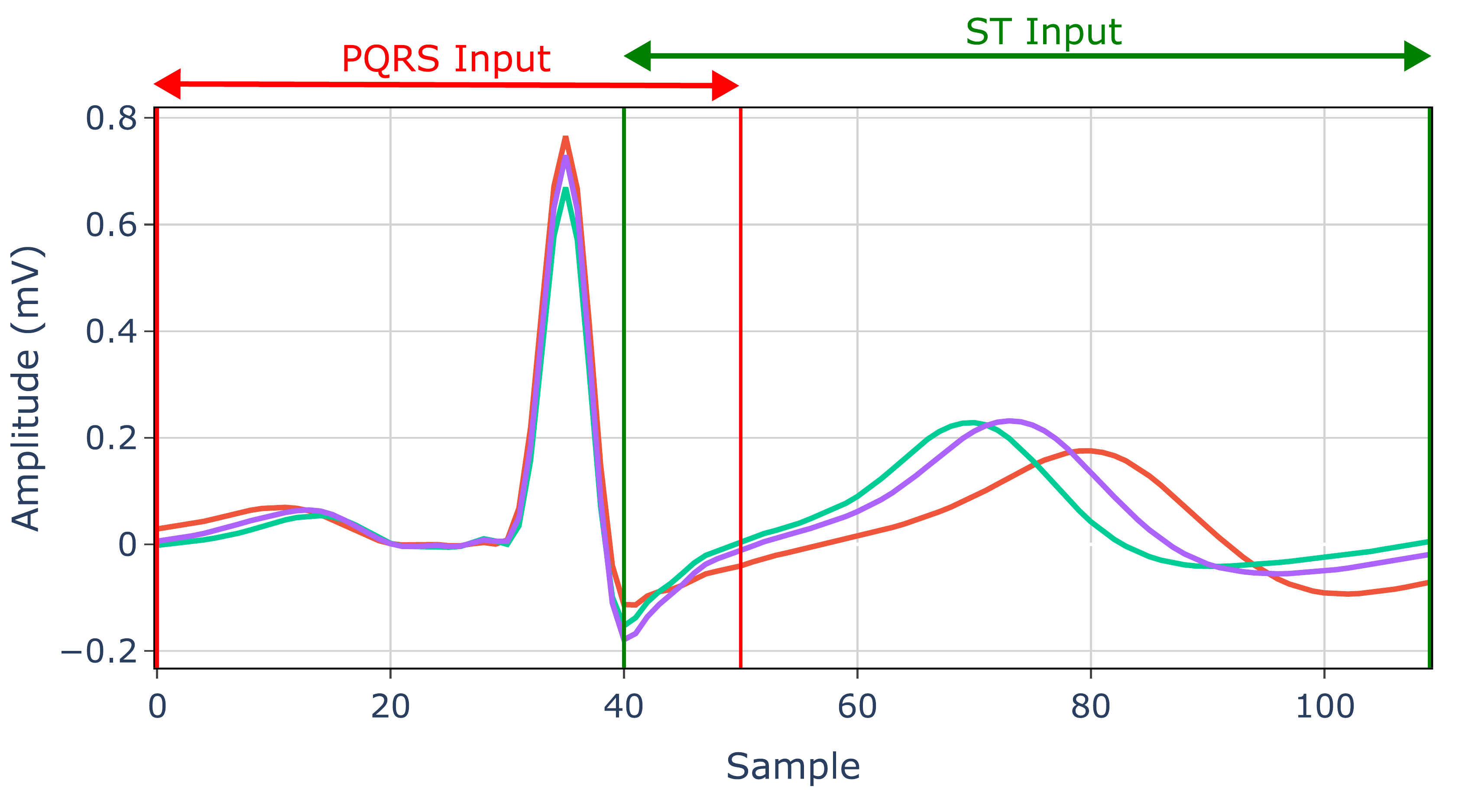}}
\caption[PQRS and ST intervals split for DE model]{The input signal is split to PQRS and ST intervals for the Dual Expert model with a 50 ms overlap.}
\label{fig:pqrs_st}
\end{figure}

CNNs are designed to learn convolutional kernels that span the entire input, yet distinct features naturally appear across different segments of the heartbeat, such as the PQRS and ST intervals, as illustrated in \figref{pqrs_st}. These differences limit the effectiveness of shared kernels across segments. Additionally, CNNs exhibit translation equivariance \cite{app10093161}, allowing them to recognize features despite minor shifts, which is valuable in capturing variations in the ST interval caused by heart rate fluctuations.
To leverage this structural knowledge of the ECG and the unique characteristics of CNNs, we propose a novel CNN-based architecture that processes the PQRS and ST intervals independently.

The Dual Expert (DE) Model consists of two Standard CNN models, each processing a different segment of the input signal. The first model processes the PQRS interval, which corresponds to the first 250 ms of the input signal, and is referred to as the PQRS model. The second model processes the ST interval, the last 350 ms of the input signal, and is referred to as the ST model.

The training of the DE model involves two training stages. In stage \uppercase\expandafter{\romannumeral 1}, each model is trained independently on the Target set data for classification. After training, the Fully Connected (FC) layers are removed, and the backbone CNN models serve as feature extractors for the DE model, as illustrated in \figref[b]{standard_dual_models}.
In stage \uppercase\expandafter{\romannumeral 2}, the features extracted by the PQRS and ST backbones are flattened and fused through concatenation into the final feature vector, which is passed into the MLP classifier. At this stage, the weights of the backbone networks are frozen, and only the FC layers of the classifier are trained.

\begin{figure}[!t]
    \myhyperlabel{standard_dual_models}
    \centering
    \includegraphics[width=\columnwidth]{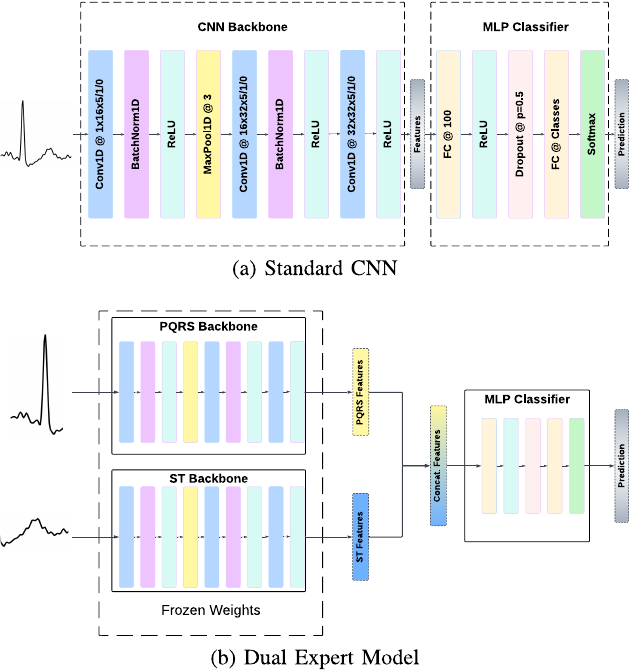}
    \caption{(a) The architecture of the Standard CNN. The CNN backbone consists of 1D convolutional layers (in\_ch$\times$out\_ch$\times$kernel/size/stride/padding),
    batch normalization, MaxPooling, and ReLU activations. The feature maps are flattened and passed to the MLP classifier, which includes Fully Connected (FC) layers, ReLU activation, dropout, and a final softmax layer for classification. (b) The DE model includes the pre-trained PQRS and ST backbones, which share the architecture with the Standard model backbone. The MLP classifiers of the Standard and DE models differ only by the input dimension.}
    \label{fig:standard_dual_models}
\end{figure}
\subsection{Domain Adaptation}
We utilize the Auxiliary set, which includes recordings under various conditions, including exercise. Since the Auxiliary data is not part of our evaluation set (Target data), its exercise data can be used for training. While augmentation accounts for ST interval variability in the time domain, it does not fully capture the variability associated with heart rate elevation, such as amplitude changes, which limits the classifier's ability to generalize to genuine exercise data.

\begin{figure}[!t]
    \myhyperlabel{domain_adaptation_clf}
    \centering
    \includegraphics[width=\columnwidth]{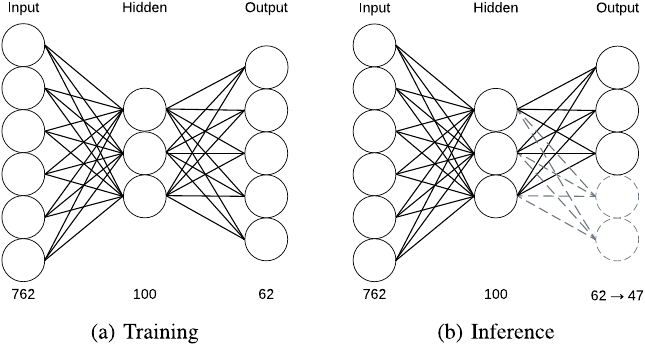}
    \caption{Classifier output dimensions during domain adaptation. (a) During training the output dimension includes subjects from Target and Auxiliary sets. (b) After training the classes corresponding to the Auxiliary set are removed.}
    \label{fig:domain_adaptation_clf}
\end{figure}

To direct the classifier toward learning condition-invariant features, in stage \uppercase\expandafter{\romannumeral 2} we train the DE model using genuine (non-augmented) data from both the Target and Auxiliary sets.
To accommodate the additional 15 subjects from the Auxiliary data, we expand the output layer of the classifier and train the model to classify all the subjects.
After training, we remove the additional neurons for inference, retaining only the classes corresponding to the subjects in our Target data as illustrated in \figref{domain_adaptation_clf}.
\section{Experiments}
\subsection{Database and Evaluation}
The University of Toronto ECG Database (UofTDB) \cite{Wahabi2014} consists of 1019 subjects, with recordings collected over a period of six months in up to five different conditions.
However, the majority of subjects participated only in the first session, which included recordings in the sitting position only. Only 100 subjects were recorded in subsequent sessions, with 55 of them having recordings in all five conditions.

The signals were recorded with dry electrodes in an off-the-person setting \cite{8392675}, captured from subjects' fingertips similar to Lead \uppercase\expandafter{\romannumeral 1} configuration at a sample frequency of 200Hz.

For our evaluation, we include the 55 subjects with samples recorded across all the conditions to ensure consistency in a multi-session setting. While our training set contains data only from sitting and standing positions, incorporating all conditions in the evaluation is important for achieving a thorough evaluation of the model's robustness and effectiveness under different conditions, including exercise, supine, and tripod.

To rigorously evaluate our approach, we preprocessed the exercise data to ensure that participants exhibited ECG responses consistent with normal physiological variations under physical stress. Atypical ECG waveforms, particularly those specific to physical stress conditions, may indicate underlying pathologies or anomalies. Since our methods leverage the natural variations in ECG waveforms, we excluded such cases from this study to maintain the validity of our findings and focus on healthy individuals.
Out of the initial 55 subjects, our analysis identified several subjects with atypical ECG patterns, including those exhibiting atypical changes exclusively during exercise. Using statistical measures detailed in the supplementary
material, we classified 8 of these subjects as outliers and excluded them from the study, resulting in a refined Target set of 47 subjects.

We use sessions S1, S2, S4, and S6 as our training data which consists of recordings in sitting and standing positions. For performance evaluation we use sit and exercise conditions from S3 along with supine and tripod positions from S5. The data split is summarized in Table~\ref{tab:session_conditions}.

Additionally, 15 subjects had exercise condition recordings but were excluded from the Target set due to missing recordings in at least one other position.
Their data was utilized to form an Auxiliary set, which is used for domain adaptation to improve the generalization of the MLP classifier.

\begin{table}[!t]
    \centering
    \caption{Train and Test Data Split}
    \label{tab:session_conditions}
    \begin{tabular}{|c|p{0.6cm}|p{0.7cm}|p{0.6cm}|p{0.7cm}|p{1.05cm}|p{1.05cm}|}
    \cline{2-7}
    \multicolumn{1}{c}{} & \multicolumn{4}{|c|}{\textbf{Train Sessions}} 
    & \multicolumn{2}{|c|}{\textbf{Test Sessions}} \\ 
    \hline
    \textbf{Session} & \textbf{S1} & \textbf{S2} & \textbf{S4} & \textbf{S6} & \textbf{S3} & \textbf{S5} \\ 
    \hline
    \textbf{Condition} & sit & sit,\newline stand &  sit & sit,\newline stand & sit,\newline exercise & supine,\newline tripod  \\ 
    \hline
    \end{tabular}
    \vspace{0.5em}
\end{table}

We evaluate our performance in the closed-set Identification setting, where we have a fixed set of $N$ subjects $S=\{s_1,s_2,...,s_n\}$.
Given an input signal $sig$, the task is to predict the identity of the corresponding subject $s_{i}\in S$.
This setting is formulated as a multi-class classification problem, where the system assigns each input to one of the pre-defined classes representing the subjects.
The evaluation metrics for identification are the Identification Rate (IDR) and False Identification Rate (FIR) also known as accuracy and error rate respectively. They are defined as follows:

\begin{equation}
    \text{IDR} = \frac{\text{No. of correct predictions}}{\text{Total No. of trials}} = \frac{1}{N} \sum_{i=1}^{N} \mathbb{I}(\hat{y}_i = y_i)
    \label{eq:ir}
\end{equation}
\begin{equation}
    \text{FIR} = \frac{\text{No. of incorrect predictions}}{\text{Total No. of trials}} = 1-\text{IDR}
    \label{eq:fir}
\end{equation}
where
\begin{description}
    \item[$N$] \qquad\enspace total number of samples;
    \item[$\hat{y}_i$] \qquad\enspace predicted label for the $i$-th sample;
    \item[$y_i$] \qquad\enspace true label for the $i$-th sample;
    \item[$\mathbb{I}(\hat{y}_i=y_i)$] \qquad\enspace indicator function that equals $1$ if $\hat{y}_i = y_i$ and \text{\qquad\enspace}$0$ otherwise.
\end{description}
Among the numerous databases used in the field, UofTDB stands out as the singular non-private database that includes subjects with both resting and exercise conditions data \cite{8392675, Wahabi2014}.
To the best of our knowledge, \cite{Jyotishi2022} is the only study to explicitly report IDR for exercise conditions from UofTDB, whereas other studies evaluating exercise data use private, non-publicly accessible databases.
Both our study and \cite{Jyotishi2022} conduct evaluations in an inter-session setting, however, the authors' approach involved training on data from the sit position only, resulting in a low IDR of 7.98\% for exercise conditions.
Our approach differs as we use multi-session training with data from both sitting and standing positions, making direct comparisons between the results of the two studies not applicable.

Therefore, due to the lack of comparable studies, we create two baseline models based on the Standard CNN architecture. One model is trained without augmentation, while the other incorporates augmentation techniques inspired by state-of-the-art methods, providing reference points for evaluating our proposed approaches. Further details on these models are provided in the following sections.

In the following experiments, we apply a sliding window of size $W=10$ when averaging heartbeats to reduce noise and improve SNR.
To ensure a fair comparison, all the methods were tested on the same data with a consistent preparation process. Additionally, the DE model architecture is based on the Standard CNN architecture, ensuring that any observed differences in performance reflect the methodological enhancements rather than fundamental architectural differences.

The train/validation data split is 80/20 respectively, with stratification by subject identity. The test data is taken from distinct sessions, as shown in Table~\ref{tab:session_conditions}.

All reported results are averaged over 10 runs to ensure reliability and consistency.
The CNN models were implemented with Pytorch\cite{paszke2019pytorchimperativestylehighperformance} and optimized with Adam optimizer\cite{kingma2017adammethodstochasticoptimization}, learning rate $10^{-3}$, cross-entropy loss, and early stopping of 20 epochs.

\subsection{Standard CNN Reference}
\label{chap:scr}
The first baseline model uses the Standard CNN architecture (\figref[a]{standard_dual_models}), and is referred to as SCR.
The SCR model is trained on the Target set data without augmentation, and its hyperparameters were tuned using the validation set.
These optimized hyperparameters were consistently applied across all subsequent experiments.

The purpose of the SCR model is to establish baseline results based on our preprocessing flow and data composition, serving as a reference point for evaluating the performance of our proposed methods.

\subsection{Augmented CNN Reference}
\label{chap:acr}
To investigate the efficacy of conventional augmentation methods, we establish the second baseline by training the SCR model with ST interval augmentation, referred to as ACR.

In line with previous studies, this augmentation follows \hyperlink{individual_augmentation}{Algorithm 1}, however, instead of calculating individual ranges (as detailed in lines \ref{lst:line:pa_start}-\ref{lst:line:pa_end} of \hyperlink{individual_augmentation}{Algorithm 1}), it uses predefined uniform ranges for all subjects.
The used predefined ranges are $T_{peak_{\text{min}}} = 25$ and $T_{peak_{\text{max}}} = 80$ sampling points relative to R-peak location, corresponding to 125 ms to 400 ms following the R-peak at a sampling rate of 200 Hz.
The training and validation data are then augmented, and the model is trained in the same manner as the SCR model.

\subsection{DE Model with Personalized Augmentation and Domain Adaptation}
\label{chap:dem_trainig}
\begin{figure}[!t]
\myhyperlabel{overall_methods}
\centering
\includegraphics[width=0.9\columnwidth]{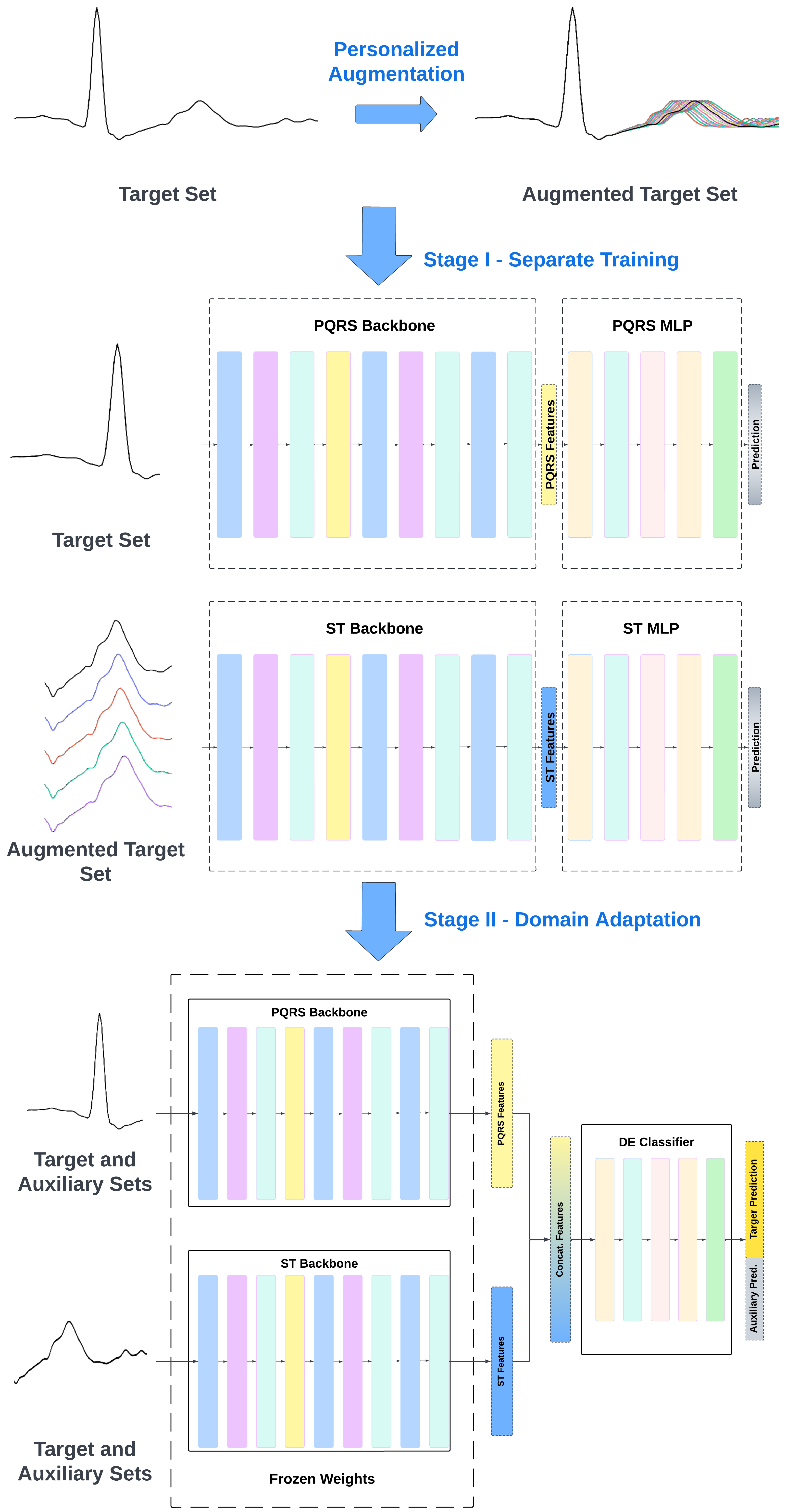}
\caption[Overview of DE-PADA training process]
{
 DE-PADA Training Process. The Target data is augmented with Personalized Augmentation, followed by separate training of the PQRS and ST models on the augmented data (Stage I). The MLP classifiers are then removed, retaining the CNN backbones with frozen weights as feature extractors. In Stage II, the DE classifier is trained using genuine data from both the target and auxiliary sets, with auxiliary classes removed from the output layer at the end.
}
\label{fig:overall_methods}
\end{figure}

The Dual Expert with Personalized Augmentation and Domain Adaptation (DE-PADA), as the name suggests, integrates the DE architecture with personalized augmentation and domain adaptation of the classifier. The training process is illustrated in \figref{overall_methods}.

First, the Target data is augmented using the proposed Personalized Augmentation, after which the PQRS and ST models are trained separately on the classification task. Following this training, the FC layers are removed, leaving only the CNN backbone networks with frozen weights to serve as feature extractors for the DE model. These backbone networks, having been trained on high-variability data through augmentation, are capable of extracting meaningful features from high heart rate data, including exercise conditions.

To enhance the classifier's ability to learn condition invariant features, we incorporate the Auxiliary set into the training process. 
This set includes an additional 15 subjects with recordings that capture both low heart rates during resting states and high heart rates following exercise, providing a comprehensive range of data for each individual in the Auxiliary set.
By expanding the training data with these diverse examples, we aim to improve the classifier's robustness and performance by leveraging the inherent similarities in how ECG signals change between different conditions across individuals.
For instance, we expect to observe some similar patterns in ECG signal changes among individuals before and after exercise.

\subsection{Results}
The dataset primarily consists of low heart rate and post-exercise records, with a noticeable lack of moderate heart rate ranges.
To address this and provide a more detailed analysis of performance at elevated heart rates, we split the exercise recordings into two distinct phases, rather than evaluating them under a single category.

\begin{figure}[!t]
\myhyperlabel{hr_change}
\centering
\includegraphics[width=\columnwidth]{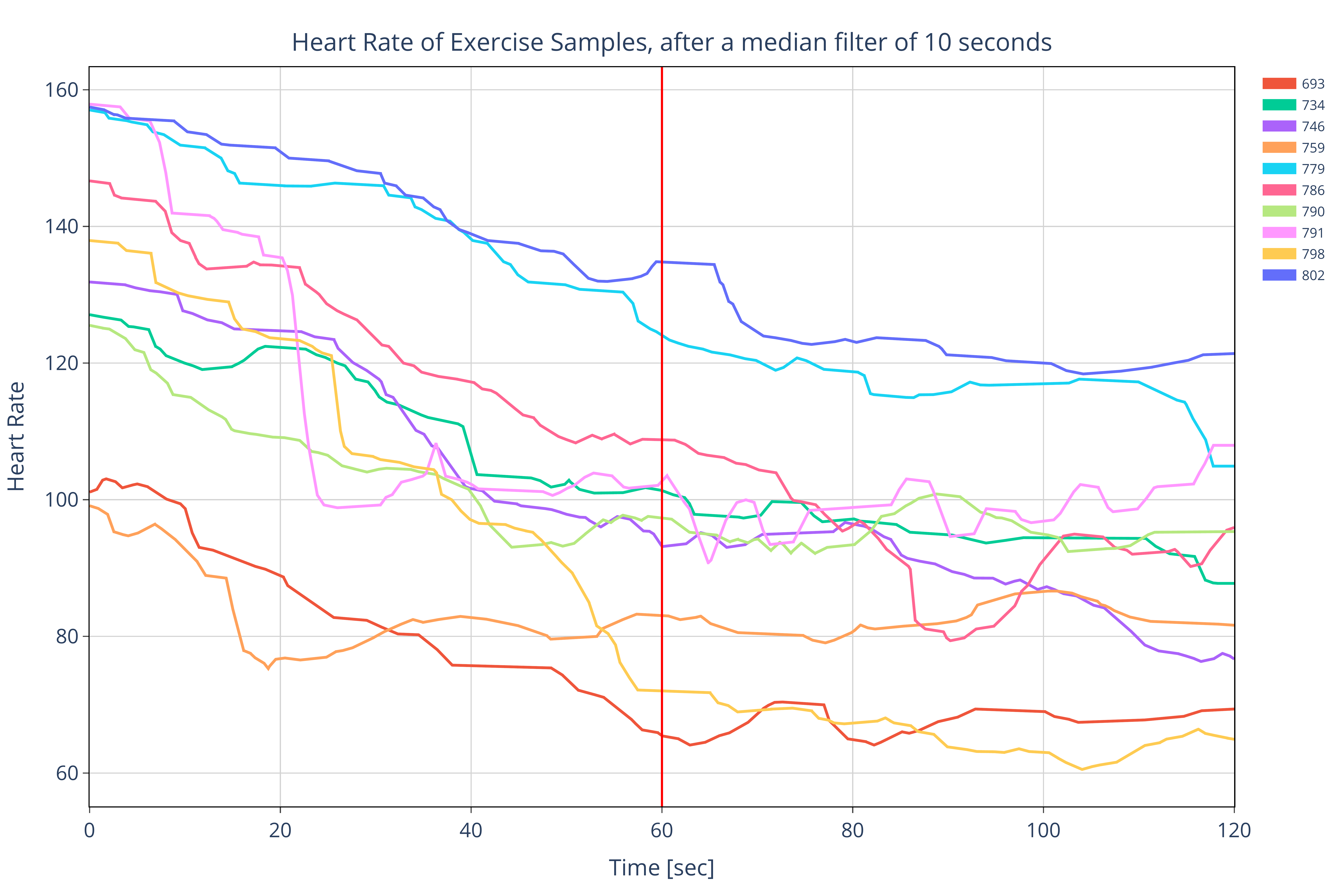}
\caption{Heart rate changes during exercise recovery, with a rapid initial decrease in the first minute, followed by a gradual stabilization in the second minute.}
\label{fig:hr_change}
\end{figure}

Each exercise recording was captured for 2 minutes, immediately following 3-5 minutes of physical activity.
As shown in \figref{hr_change}, the first minute of the recording represents the Initial Recovery Phase, characterized by a high heart rate and a rapid decline as the body begins to recover from physical stress. The second minute represents the Late Recovery Phase, where the heart rate is moderate and mostly stable. Although the duration of the initial recovery may vary between subjects, we uniformly set it as one minute for consistency in our analysis.
Separating these phases allows us to analyze identification performance more effectively across different heart rate levels and variability, especially given the scarcity of moderate heart rate samples in the dataset.

We will refer to the initial and late recovery phases as Ex\_P1 and Ex\_P2, respectively.

\subsubsection{Sit and Exercise Performance}
\label{chap:sit_ex_results}
In this section, we present the results for the sit position, the most commonly studied and prevalent resting condition in real-life scenarios, and for exercise, which is the primary focus of this work.

The SCR model, as shown in Table~\ref{tab:aug_baseline_performance}, achieved high performance for the sit position with an Identification Rate (IDR) of 97.76\%, demonstrating its effectiveness in low-variability, resting state data.
However, its performance drops significantly in exercise conditions, with IDR of 77.38\% in Ex\_P2 and 54.40\% in Ex\_P1.
This decline underscores the limitations of the SCR model in handling high-variability scenarios typical of elevated heart rates and exercise, with a more severe decline at higher heart rates.

\begin{figure}[!t]
\myhyperlabel{sit_error_final}
\centering
\includegraphics[width=0.9\columnwidth]{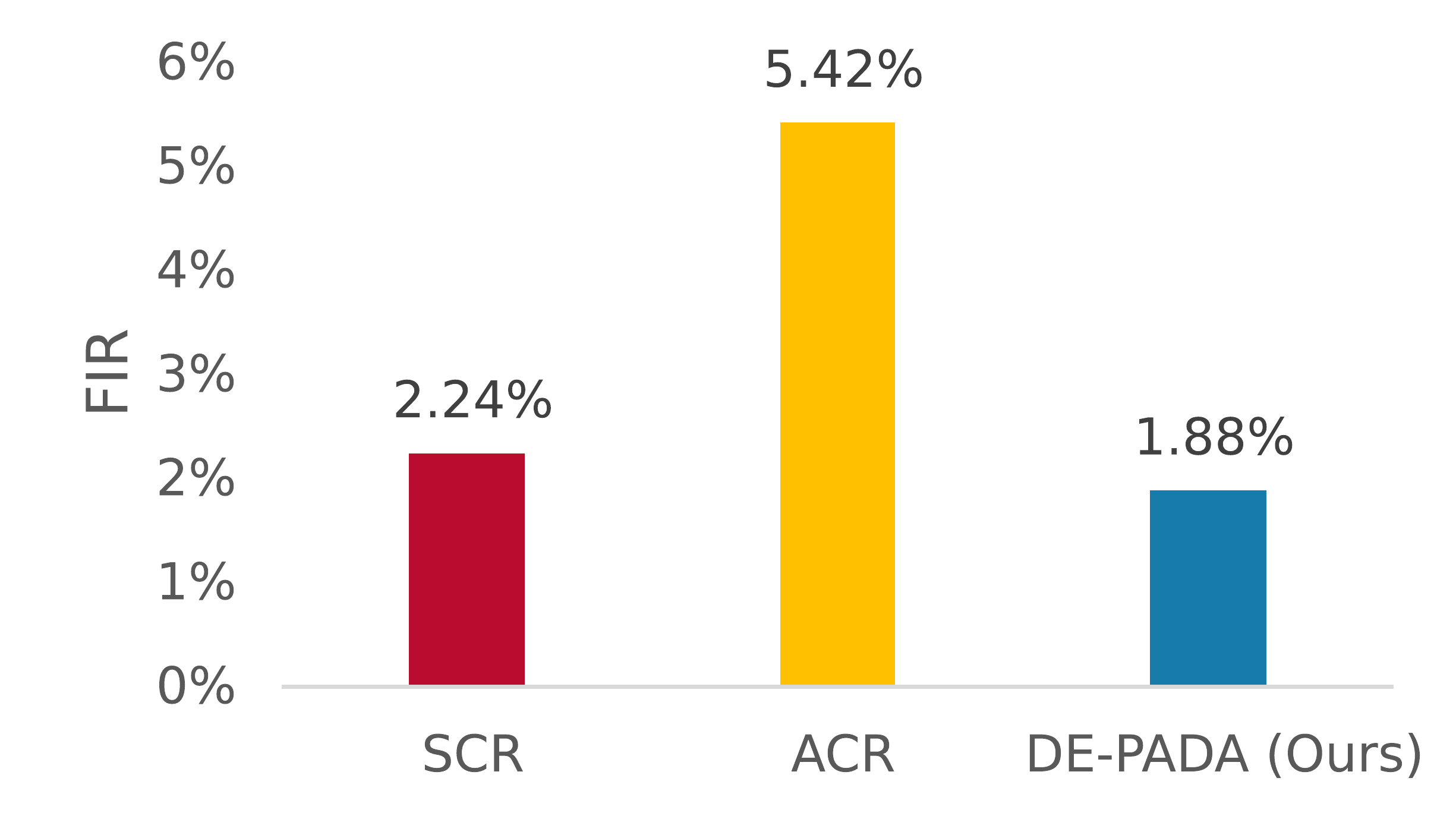}
\caption[FIR on sit position]{The FIR results for the sit position more than doubled with the use of conventional augmentation. The DE-PADA not only maintains the performance in the sit position but also achieves the lowest FIR.}
\label{fig:sit_error_final}
\end{figure}

To improve performance under exercise conditions, the ACR model employs conventional augmentation.
This approach resulted in improved IDR of 81.15\% and 66.63\% for Ex\_P2 and Ex\_P1, respectively, demonstrating the benefits of augmentation in addressing the increased variability of exercise data.
However, this improvement came at the cost of reduced performance for sit position, where the IDR decreased to 94.58\%.
As displayed in \figref{sit_error_final}, the False Identification Rate (FIR) for the sit position more than doubled, highlighting a significant trade-off associated with applying conventional augmentation methods without accounting for individual subject characteristics.

While this trade-off may initially seem acceptable, such a decrease in performance for the most common and fundamental position has a substantial impact on the overall usability of the system.

The proposed DE-PADA model demonstrates superior performance under exercise conditions, achieving IDR of 86.45\% in Ex\_P2 and 68.95\% in Ex\_P1, outperforming both baseline models.
Most importantly, this improvement in exercise performance does not come at the cost of performance for the sit position.
The DE-PADA model achieved a marginally improved IDR of 98.12\% on the sit position, surpassing both baselines and ensuring robust identification across both resting and active states.
These results emphasize the model’s capability to effectively manage high-variability data while maintaining high performance for sit position.

\begin{table}[!t]
    \centering
    \caption[Performance on sit position and exercise conditions]{Sit and exercise IDR compared to the baseline models.}
    \label{tab:aug_baseline_performance}
    \begin{tabular}{lccc}
        \hline
        Method                   & Sit              & Exercise Phase 2  & Exercise Phase 1  \\
        \hline
        SCR (\ref{chap:scr})                     & 97.76\%          & 77.38\%           & 54.40\%           \\
        ACR (\ref{chap:acr})                     & 94.58\%          & 81.15\%           & 66.63\%           \\
        \textbf{DE-PADA (Ours)}  & \textbf{98.12\%} & \textbf{86.45\%}  & \textbf{68.95\%}  \\
        \hline
    \end{tabular}%
\end{table}

\subsubsection{Overall Performance}
The overall performance across all conditions, including sit, exercise phases, supine, and tripod, provides a comprehensive evaluation of each model's robustness and generalizability. As shown in Table~\ref{tab:final_performance}, the DE-PADA model consistently outperforms the baseline models across all conditions, demonstrating its effectiveness in handling the variability inherent in diverse postures and heart rate levels.

For the supine position, the DE-PADA model achieved an IDR of 96.08\%, surpassing both the SCR (94.81\%) and ACR (93.73\%) baseline models. This indicates the DE-PADA model's capability to maintain high performance even in less frequently encountered conditions. In the tripod position, DE-PADA also outperformed with an IDR of 87.34\%, compared to 85.77\% for SCR and 82.71\% for ACR, highlighting its resilience in scenarios with altered body positioning, which often pose challenges due to changes in ECG signal morphology.

This comprehensive improvement underscores the strength of the proposed DE-PADA model in maintaining superior performance in exercise phases while not compromising on any resting positions. DE-PADA demonstrates its potential for real-world applications, where biometric systems must reliably operate across diverse conditions and activities.
\begin{table}[!t]
    \centering
    \caption{Overall IDR results for all the conditions.}
    \label{tab:final_performance}
    \resizebox{\columnwidth}{!}{%
        \begin{tabular}{lccccc}
            \hline
            Method                  & Sit              & Exercise         & Exercise         & Supine           & Tripod           \\
                                    &                  & Phase 2          & Phase 1          &                  &                  \\
            \hline
            SCR                     & 97.76\%          & 77.38\%          & 54.40\%          & 94.81\%          & 85.77\%          \\
            ACR                     & 94.58\%          & 81.15\%          & 66.63\%          & 93.73\%          & 82.71\%          \\
            \textbf{DE-PADA (Ours)} & \textbf{98.12\%} & \textbf{86.45\%} & \textbf{68.95\%} & \textbf{96.08\%} & \textbf{87.34\%} \\
            \hline
        \end{tabular}
    }
\end{table}
\section{Ablation Study}
\label{sec:ablation_study}
In this section, we analyze the contribution of each component of the proposed DE-PADA model.
To achieve this, we perform an ablation study by systematically removing individual components from the final model to assess their impact on the results for both the sit and exercise phases.

As a reminder, the training of the DE-PADA model consists of two stages: in Stage \uppercase\expandafter{\romannumeral 1}, the backbone PQRS and ST models were trained separately on the Target set with personalized augmentation to optimize feature extraction; in Stage \uppercase\expandafter{\romannumeral 2}, the classifier was trained with domain adaptation on genuine, non-augmented data to enhance its ability to learn invariant features across different conditions.

The ablation study is conducted in two primary scenarios. In the first scenario, we evaluate our approach by training the classifier only with domain adaptation, without augmentation, consistent with the methodology used up to this point.
In the second scenario, we examine the results when the classifier is trained with both domain adaptation and personalized augmentation. The augmentation was limited to the Target set since the Auxiliary set has partial representation across sessions and lacks the required data for T-peak range calculations, which are integral to our personalized augmentation method. It's important to note that all the evaluations were conducted on the Target set, consistently with previous evaluations, with the reported IDR being the average of 10 runs.

The results are reported for four different cases, to isolate the effects of each component:
\begin{itemize}
    \item \textbf{DE-PADA}: The complete DE-PADA model, trained with all proposed components, serves as the reference for comparison.
    \item \textbf{DE-PADA\textbackslash DE}: The Standard CNN model is used instead of the DE model, removing the benefit of handling different segments separately.
    \item \textbf{DE-PADA\textbackslash PA}: The DE-PADA model was trained without personalized augmentation at any stage.
    \item \textbf{DE-PADA\textbackslash DA}: The DE-PADA model was trained without using domain adaptation.
\end{itemize}

\subsection{Non-Augmented Classifier}
The results of the ablation study, when the classifier is trained without augmentation, are presented in Table~\ref{tab:ablation_notaugmented}.
This table illustrates the impact of removing various components from the DE-PADA model without using augmented data in the classifier's training stage.

Comparing the DE-PADA\textbackslash DE model to the full DE-PADA model, we observe a significant decrease in performance for the sit position (95.55\% vs. 98.12\%), along with a slight increase in performance for Ex\_P1 (69.59\% vs. 68.95\%).
This suggests that the DE model is particularly effective at preserving performance in low-variability scenarios, such as the sit position.
In high-variability conditions, the DE model offers only marginal improvements in Ex\_P2 and a slight reduction in Ex\_P1.
These findings highlight the advantage of processing different ECG segments separately in the DE model, which is particularly valuable for maintaining robustness across various conditions.

For the DE-PADA\textbackslash PA model, where training was conducted without personalized augmentation, the highest performance is observed in the sit condition (98.52\%), slightly outperforming the full DE-PADA model (98.12\%). This aligns with the presented results on conventional augmentation, which suggest that augmentation in general can negatively impact performance in scenarios with sufficient training data and low variability. However, in this case, the reduction in performance is relatively small. Notably, the absence of personalized augmentation causes a substantial decline in performance for exercise phases, with Ex\_P2 at 81.34\% and Ex\_P1 at 59.18\%. These results underscore the importance of personalized augmentation in managing heart rate variability, as its exclusion significantly compromises the model's effectiveness in high-variability settings.

The DE-PADA\textbackslash DA model, which was trained without domain adaptation, also shows reduced performance compared to the full DE-PADA model, particularly in Ex\_P2 (80.70\%) and Ex\_P1 (59.48\%), with a minor decrease in the sit condition (97.94\%). This indicates that domain adaptation, similarly to augmentation, is crucial for enhancing the model's generalization ability, especially in conditions involving elevated heart rates and varying postures.
\begin{table}[!t]
    \centering
    \caption{Ablation study results when the classifier is trained without augmentation.}
    \label{tab:ablation_notaugmented}
    \begin{tabular}{lccc}
        \hline
        Method                   & Sit              & Exercise Phase 2  & Exercise Phase 1  \\
        \hline
        DE-PADA                  & \textbf{98.12}\% & \textbf{86.45\%} & 68.95\%          \\
        DE-PADA\textbackslash DE & 95.55\%          & 84.94\%          & \textbf{69.59\%} \\
        DE-PADA\textbackslash PA & \textbf{98.52\%} & 81.34\%          & 59.18\%          \\
        DE-PADA\textbackslash DA & 97.94\%          & 80.70\%          & 59.48\%          \\
        \hline
    \end{tabular}
\end{table}
\subsection{Augmented Classifier}
In the second scenario, the classifier was trained with both domain adaptation and personalized augmentation, with the augmentation limited to the Target set as mentioned above.

The results presented in Table~\ref{tab:ablation_augmented} show that the full DE-PADA model achieves the highest performance for exercise phases, with IDR values of 86.28\% for Ex\_P2 and 71.78\% for Ex\_P1. This demonstrates the effectiveness of combining domain adaptation with personalized augmentation in managing high-variability data. The model also surpasses the non-augmented variant from the previous section in Ex\_P1 due to the inclusion of augmentation for the classifier, and the generation of synthetic elevated heart rate examples. However, the decrease in performance for the sit condition reflects the trade-off discussed in Subsection~\ref{chap:sit_ex_results}.

The importance of the DE model is further highlighted in this scenario, as the DE-PADA\textbackslash DE model exhibits a decline in performance across all conditions.
Similarly, the model trained without domain adaptation shows reduced performance, although the reduction in exercise phases is less pronounced than in the non-augmented scenario, as the inclusion of augmentation helps the classifier handle exercise phases more effectively.

Notably, in all models where Stage \uppercase\expandafter{\romannumeral2} included augmentation, performance on the sit position decreased. The DE-PADA\textbackslash PA model, identical to the one in the previous section, achieves the highest performance on the sit position.
\begin{table}[!t]
    \centering
    \caption{Ablation study results when the classifier is trained with personalized augmentation.}
    \label{tab:ablation_augmented}
    \begin{tabular}{lccc}
        \hline
        Method                   & Sit              & Exercise Phase 2  & Exercise Phase 1  \\
        \hline
        DE-PADA                  & 96.39\%          & \textbf{86.28\%} & \textbf{71.78\%} \\
        DE-PADA\textbackslash DE & 95.37\%          & 82.57\%          & 67.09\%          \\
        DE-PADA\textbackslash PA & \textbf{98.52\%} & 81.34\%          & 59.18\%          \\
        DE-PADA\textbackslash DA & 96.33\%          & 84.32\%          & 65.33\%          \\
        \hline
    \end{tabular}
\end{table}

We hypothesize that the partial accounting for heart rate changes introduced through augmentation is particularly beneficial in scenarios with large performance gaps, as observed in related studies that use augmentation in low-data-availability settings. This explains the observed improvement in exercise performance.
However, in cases where the training set contains adequate amounts of representative data, achieving a high initial IDR, the augmentation's inability to fully replicate the authentic changes in the ECG waveform may hinder performance.
\section{Augmentation Effect on Feature Space}
To gain deeper insights into the impact of augmentation on our models and whether ST interval normalization remains relevant when training the model with augmentation, we conducted a detailed analysis of the ST model feature space. The ST model, which is the component of DE-PADA responsible for extracting ST interval features (\figref[b]{standard_dual_models}), is the only part affected by the augmentation, therefore we can ignore the PQRS model features in this analysis.
The analysis was performed on the test set, to analyze exercise data and avoid any bias from the training process.\\
The analysis involved the following steps:

\begin{enumerate}
    \item \textbf{Normalization of ST Interval}: We began by normalizing the ST interval for the test data similarly to \cite{Hwang2021}, but accordingly to each individual’s specific fit, rather than a global fit. First, the duration of the ST interval is calculated from the linear fit at the average heart rate of the subject's training data. Then, the T-wave of all the data corresponding to the subject is resampled to the calculated duration.

    \item \textbf{Feature Extraction with Non-Augmented ST Model}: We extracted features from both the original and normalized test data using an ST model that was trained without any augmentation.

    \item \textbf{Feature Extraction with Augmented ST Model}: Next, we repeated the feature extraction process using an ST model that was trained with personalized augmentation.

    \item \textbf{Dimensionality Reduction with t-SNE}: To visualize the extracted features, we applied t-SNE for dimensionality reduction \cite{vandermaaten08a}. This technique allows us to explore the clustering behavior and the distribution of the features in a two-dimensional space.
    Since t-SNE is a stochastic iterative algorithm, which can result in a different reduction on each run, we grouped the features resulting from each model and applied t-SNE to each group.

    \item \textbf{Comparison of Normalization Effects}: Finally, we compared the effects of normalization on each model's feature space and how these effects differ between the two models.
\end{enumerate}

\begin{figure}[!t]
    \myhyperlabel{tsne_graphs}
    \centering
    \includegraphics[width=\columnwidth]{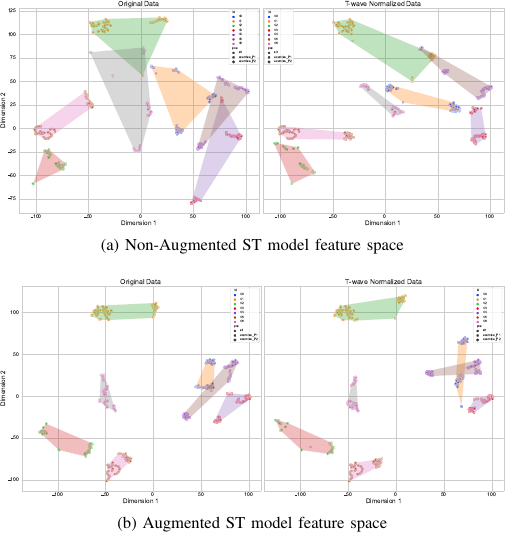}
    \caption[t-SNE 2D Feature Reduction of Original and Normalized Data]
        {
            t-SNE 2D Feature Reduction of Original and Normalized Data.
            (a) Heartbeat normalization improves feature compactness, and the feature mapping changes considerably after normalization.
            (b) Features are compact prior to normalization, and the feature space is hardly affected by it.
        }
    \label{fig:tsne_graphs}
\end{figure}

\figref{tsne_graphs} presents the features after t-SNE dimensionality reduction for the last eight subjects in the Target set. The shape of the data points represents the condition of each sample, while the colors distinguish different subjects. 
The background color is a convex hull that groups all data points of each subject; it serves as a visual aid and does not correspond to the classifier's decision boundaries.
\subsection{Non-Augmented ST Model}
\figref[a]{tsne_graphs} illustrates the feature space of the non-augmented model. On the left side, representing the original data, it is evident that most subjects form more than two clusters in the feature space, with subject 40 (blue) displaying four to five clusters. However, the features of the normalized data show a reduction in the number of clusters for each subject, indicating that normalization leads to a more compact feature space.

\subsection{Augmented ST Model}
Upon examining the features of the original data in \figref[b]{tsne_graphs}, it is observed that the feature space is initially compact, similar to the compactness seen in the normalized data features of the non-augmented model.

Since each model underwent a separate dimensionality reduction, a direct comparison of the data point locations between (a) and (b) is not possible. However, the changes in locations between the left and right sides of both models can be compared.
For the non-augmented model, most subjects show a noticeable shift in the location of their data points after normalization. However, for the augmented model, the feature space exhibits minimal changes for most subjects, suggesting that the model's feature space is robust to T-wave variability, as it is largely unaffected by T-wave normalization.

\subsection{Observations}
Our experiments using data normalization in conjunction with augmentation did not provide any additional benefit, which aligns with the observations in \figref[b]{tsne_graphs}.
Furthermore, visualizing the feature spaces of both models reveals that even at the ST interval level, neither augmentation nor normalization fully compensates for the changes occurring in Ex\_P1 and Ex\_P2.
Despite t-SNE being an unsupervised dimensionality reduction algorithm, sit and exercise data points were clustered separately, with each subject having at least two clusters.

\section{Conclusions}
In this paper, we addressed the challenge of ECG-based user identification across varying body postures and physiological states, particularly under post-exercise conditions with elevated heart rates. We proposed a comprehensive approach that combines a novel Dual Expert (DE) model with Personalized Augmentation and Domain Adaptation (DE-PADA) to effectively handle the intra-subject variability of ECG signals across diverse conditions. Each of these three components leverages the morphological characteristics of ECG signals to achieve robust identification performance, and combined, they significantly surpassed the reference models in all tested scenarios.

We proposed a Dual Expert (DE) architecture that separately attended to the PQRS and ST intervals, effectively preserving performance in resting states. We introduced a Personalized Augmentation algorithm that augments the ST interval within predicted subject-specific ranges, significantly improving identification under exercise conditions. Additionally, we presented a domain adaptation variant that utilizes data from additional subjects with both resting and active state data. This approach enabled the classifier to learn patterns common to the population including the Target set subjects, thereby enhancing its generalization ability.

The DE-PADA model consistently outperformed the baseline models across all tested conditions. It achieved notable improvements in identification rates, increasing from 77.38\% to 86.45\% for Exercise Phase 2 and from 54.4\% to 68.95\% for Exercise Phase 1 compared to the standard reference model. In addition, the DE-PADA model maintained high accuracy in stable resting conditions such as sitting, achieving an identification rate of 98.12\%, which not only countered the reduction observed in the augmented reference model but also surpassed the baseline performance.

Furthermore, we analyzed the effect of personalized augmentation on the feature space of the ST model and demonstrated its effectiveness in reducing some of the intra-subject variability and creating a more compact feature space. However, after t-SNE dimensionality reduction, it remains evident that features from the sitting position and exercise phases are still clustered separately, indicating that additional methods, such as the proposed domain adaptation, can further reduce this gap.

\bibliographystyle{IEEEtran}
\bibliography{references}

\end{document}